\documentclass{article}

\usepackage{arxiv}

\usepackage{microtype}
\usepackage{graphicx}
\usepackage{subcaption}
\usepackage{booktabs} 
\usepackage{array,booktabs,longtable,tabularx,tabulary}
\usepackage{multicol, multirow}
\usepackage{amsmath,amsfonts,amssymb}

\usepackage{hyperref}

\graphicspath{ {./FIGURES/}}

\usepackage{changes}
\usepackage{array}
\newcolumntype{P}[1]{>{\centering\arraybackslash}p{#1}}
\newcolumntype{M}[1]{>{\centering\arraybackslash}m{#1}}
\usepackage{tikz}
\usepackage{bm}
\usepackage[para]{threeparttable}

 \usepackage[style=ieee, sorting=none, backend=biber, doi=false, isbn=false, maxbibnames=4
  ]{biblatex}

\title{Interpretable Deep Learning for the Remote Characterisation of Ambulation in Multiple Sclerosis using Smartphones}

\author{
    Andrew P.~Creagh\thanks{Corresponding author: andrew.creagh@eng.ox.ac.uk;} \\
    Institute of Biomedical Engineering\\
    University of Oxford, UK\\
    \texttt{andrew.creagh@eng.ox.ac.uk} \\
    \And
    Florian ~Lipsmeier\\
    Roche Innovation Center \\
    Basel, CH \\
    \texttt{florian.lipsmeier@roche.com}\\
       \And
    Michael ~Lindemann\thanks{Shared last authorship;}\\
    Roche Innovation Center \\
    Basel, CH \\
    \texttt{michael.lindemann@roche.com}\\
    \And
    Maarten ~De~Vos\footnotemark[2]\\
    Department of Electrical Engineering\\
    KU Leuven, BE\\
    \texttt{maarten.devos@kuleuven.be}
}

\begin{document}

\maketitle

\begin{abstract}
The emergence of digital technologies such as smartphones in healthcare applications have demonstrated the possibility of developing rich, continuous, and objective measures of multiple sclerosis (MS) disability that can be administered remotely and out-of-clinic.
Deep Convolutional Neural Networks (DCNN) may capture a richer representation of healthy and MS-related ambulatory characteristics from the raw smartphone-based inertial sensor data than standard feature-based methodologies.
To overcome the typical limitations associated with remotely generated health data, such as low subject numbers, sparsity and heterogeneous data, a transfer learning (TL) model from similar large open-source datasets was proposed. Our TL framework leveraged the ambulatory information learned on human activity recognition (HAR) tasks collected from wearable smartphone sensor data. It was demonstrated that fine-tuning TL DCNN HAR models towards MS disease recognition tasks outperformed previous Support Vector Machine (SVM) feature-based methods, as well as DCNN models trained end-to-end, by upwards of 8--15\%. 
A lack of transparency of ``black-box'' deep networks remains one of the largest stumbling blocks to the wider acceptance of deep learning for clinical applications. Ensuing work therefore aimed to visualise DCNN decisions attributed by relevance heatmaps using Layer-Wise Relevance Propagation (LRP). Through the LRP framework, the patterns captured from smartphone-based inertial sensor data that were reflective of those who are healthy versus persons with MS (PwMS) could begin to be established and understood. Interpretations suggested that cadence-based measures, gait speed, and ambulation-related signal perturbations were distinct characteristics that distinguished MS disability from healthy participants. 
Robust and interpretable outcomes, generated from high-frequency out-of-clinic assessments, could greatly augment the current in-clinic assessment picture for PwMS, to inform better disease management techniques, and enable the development of better therapeutic interventions.
\end{abstract}

\keywords{{Gait \and deep learning \and multiple sclerosis \and digital biomarkers \and smartphones}}

\section{Introduction}
Digital health technology assessments may enable a deeper characterisation of the symptoms of neurodegenerative diseases from at-home environments \cite{RN1134}. A wealth of recent work is focusing on how digital outcomes can be captured from sensor data collected with consumer devices to represent impairment in neurodegenerative and autoimmune diseases such as multiple sclerosis (MS) \cite{RN940, RN988},  Parkinson's disease (PD) \cite{RN818, RN989}, and rheumatoid arthritis (RA) \cite{RN1206}, both remotely and longitudinally. 
MS is a heterogeneous and highly mutable disease, where people with MS (PwMS) can experience symptomatic episodes (a relapse) which fluctuate periodically and impairment tends to increase over time \cite{RN20}. Objective and frequent monitoring of the manifestations of PwMS disability is therefore of considerable importance; digital sensor-based assessments may be more accurate than conventional clinical outcomes recorded at infrequent visits in detecting subtle progressive sub-clinical changes or long-term disability in PwMS \cite{RN34}. Furthermore, earlier identification of changes in PwMS impairment are important to identify and provide better therapeutic strategies \cite{RN905}. \par
Alterations during ambulation (gait) due to MS are a amongst the most common indication of MS impairment \cite{RN289, RN294, RN293, RN710, RN728, RN290}. PwMS can display postural instability \cite{RN294}, gait variability \cite{RN293,RN710, RN728} and fatigue \cite{RN290} during various stages of disease progression. \par
The gold-standard assessment of disability in MS is the Expanded Disability Status Scale (EDSS) \cite{RN16}, as well as specific functional domain assessments such as the Timed 25-Foot Walk (T25FW), which is part of the Multiple Sclerosis Functional Composite score \cite{RN686,RN685}, and the Two-Minute Walk Test (2MWT) which also assesses physical gait function and fatigue in PwMS \cite{RN924}. \par
Body worn inertial sensor-based measurements have been proposed as objective methods to characterise gait function in PwMS \cite{RN728, RN727, RN710}. This study builds upon our previous investigations \cite{RN988}, where we have shown how inertial sensors contained within consumer-based smartphones and smartwatches can be used to characterise gait impairments in PwMS from a remotely administered Two-Minute Walk Test (2MWT) \cite{RN924}.
We have demonstrated how inertial sensor-based features can be extracted from these consumer devices to develop machine learning (ML) models that can distinguish MS disability from healthy participants. \par
These approaches, however, are constrained transformations and approximations of ambulatory function which are based on prior assumptions. 
Hand-crafted gait features are often established signal-processing metrics re-purposed as surrogates to represent aspects of PwMS gait; for instance, extracting the variance in a sensor signal in an attempt to capture gait variability in PwMS. As such, there may be greater power in allowing an algorithm to learn its own features, termed \textit{representation learning} \cite{RN824}.
Deep learning is an overarching term given to \textit{representation learning}, where multiple levels of representation are obtained through the combination of a number of stacked (hence \textit{deep}) non-linear model layers. 
Deep learning models typically describe convolutional neural networks (CNN), deep neural networks (DNN), and combined fully-connected deep convolutional neural networks (DCNN) architectures \cite{RN824}. Other architectures include recurrent neural networks (RNN), such as Long Short Term Memory (LSTM) networks, which are especially adept at modelling sequential time-series data \cite{RN824}. While CNNs are omnipresent in image recognition-based tasks, these models are often extremely successful at many time-series related tasks \cite{RN793, RN735}. For example, CNNs have been shown to act as feature extractors capable of learning temporal and spatio-temporal information directly from the raw time-series signals \cite{RN793}. The features extracted by convolutional layers can then be arranged to create a final output through fully connected (dense) layers. 
It should be noted that although there is no fundamental difference between the nomenclature `CNN' and `DCNN', in this manuscript we explicitly refer to the entire model as a DCNN in order to facilitate the distinguishment between the role of the feature extraction CNN layers and classification fully connected layers.\par
Recently, deep networks are also being applied towards inertial sensor data for a range of various activity related tasks. For instance, by far the most popular ---and most accurate--- techniques which have been applied to Human Activity Recognition (HAR) based sensor activities incorporate deep networks \cite{RN762, RN735, RN257}. Many studies are beginning to explore disease classification and symptom monitoring with wearable-generated inertial sensor data using deep networks. Representations learned using DCNNs from wearable and smartphone inertial sensors have been shown to predict gait impairment in Parkinson's disease \cite{RN989, RN770}, to predict falls risk in both the elderly \cite{RN753} and in PwMS \cite{RN264}, as well as DCNNs for subject identification tasks \cite{RN609, RN991}.

\subsection{Deep Transfer Learning for Remote Disease Classification}\label{sec:Transfer Learning}
The work presented in this study compares the performance of CNN extracted features and DCNN models against hand-crafted features previously introduced in \cite{RN988} to directly classify healthy controls (HC) and subgroups of PwMS. Despite the possibility of significant performance improvements compared to hand-crafted feature-based methods, deep networks require much more training data to make successful, robust and generaliseable decisions \cite{RN824}. Transfer learning (TL) is a machine learning technique which aims to overcome these challenges by transferring information learned between related source and target domains \cite{RN824, RN788}.
While the data may be in different domains, or the distributions may differ from the target and source tasks, transfer learning assumes that the knowledge that is learned in another task and dataset will be useful and related to the new target task. \par
Transfer learning has been successfully implemented in many computer vision tasks \cite{RN756} and for time series classification tasks \cite{RN788}, such as EEG sleep-staging \cite{RN842,RN844}, and importantly, towards accelerometery based falls prediction \cite{RN753} and within physical activity recognition \cite{RN870, RN938}. \par
\begin{figure}[t!]
	\centering
  \includegraphics[width=1\linewidth]{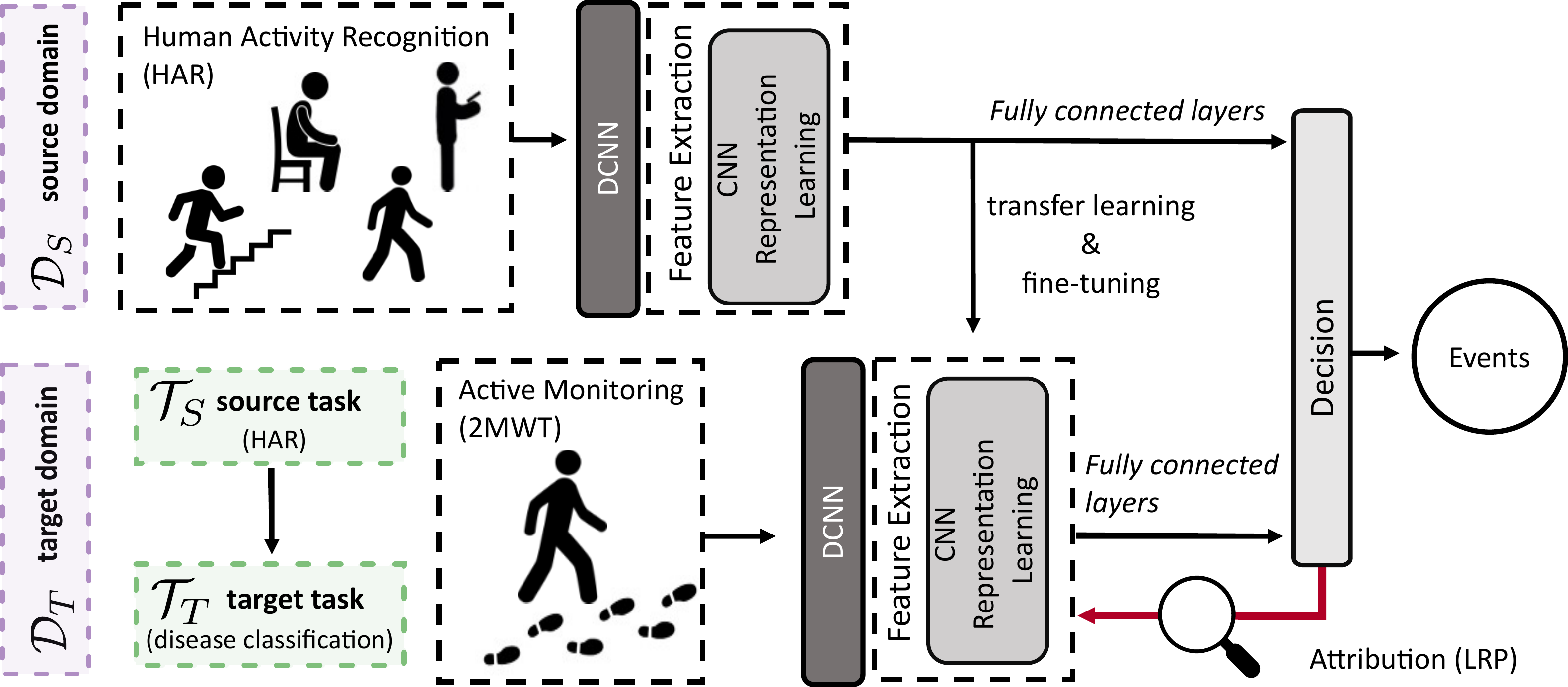}
\caption{\footnotesize \textbf{Schematic of proposed smartphone-based remote disease classification approach}. First, open-source datasets ($\mathcal{D}_S$) were utilised to learn a HAR classification task ($\mathcal{T}_S$) with a Deep Convolutional Neural Network (DCNN). Learned activity information was then subsequently transferred using the transfer learning (TL) framework, where a portion of the DCNN model is retrained on the FL datatset ($\mathcal{D}_T$), and parameters are fine-tuned towards the application of a disease recognition task ($\mathcal{T}_T$). DCNN model decisions can then be visually interpreted using attribution techniques, such as layer-wise relevance propagation (LRP), which aim to map the patterns of an input signal that are responsible for the activations within a network, and hence uncover pertinent MS disease-related ambulatory characteristics.}\label{fig:tl_overview_schematic} 
\end{figure}

We therefore aim to utilise transfer learning to supplement our model's ability to discriminate between healthy and diseased subjects in the FLOODLIGHT proof-of-concept (FL) dataset (see table \ref{table:demographics}) \cite{RN988, RN891}. Deep transfer learning was performed by first identifying relevant large (open-) source datasets from which information can be exploited. The similarity between some HAR datasets and FL, the applicability of the HAR domain task (which includes walking bouts), as well as the trove of established HAR deep network architectures, suggests that HAR may be a suitable candidate to transfer domain knowledge. We identified two publicly available datasets, UCI HAR \cite{RN847} and WISDM \cite{RN840}, which use comparable smartphone and smartwatch devices, and device affixing locations similar to that of FL. Figure \ref{fig:tl_overview_schematic} schematically illustrates the transfer learning approach undertaken, where the information learned from a HAR classification task ($\mathcal{T}_S$) and dataset ($\mathcal{D}_S$) are transferred towards a disease recognition task ($\mathcal{T}_T$) within the FL dataset ($\mathcal{T}_T$). Demographic details of the UCI HAR and WISDM HAR datasets explored in this study can be found in the accompanying supplementary material.  

\subsection{Visually Interpreting Smartphone-based Remote Sensor Models through Attribution}\label{sec:Visualisation}
Deep networks can be highly non-linear and complex, leading to an inherent difficulty in interpreting the decisions that lead to a prediction \cite{RN857, RN855}. As such, there is a considerable interest in explaining and understanding these ``black-box'' algorithms \cite{RN948}; model transparency is particularly considered a hindrance to the widespread uptake and acceptance of deep networks in medical contexts, versus less powerful, but interpretable linear models \cite{RN859}. A number of techniques have been developed in recent years to help explain deep neural networks  \cite{RN857, RN835,RN860, RN944,RN948}. Layer-wise relevance propagation (LRP) is a backward propagation technique which has gained considerable notoriety as a method to explain and interpret deep networks beyond many existing techniques \cite{RN861, RN851}. Layer-wise relevance propagation has demonstrated clinical utility in interpreting relevant parts of the brain responsible for the predictions of Alzheimer's disease (AD) \cite{RN826} and MS using MRI images \cite{RN943}. Attribution through LRP has also been successfully applied to clinical time series data such as EEG trial classification for brain–computer interfacing \cite{RN947} and, crucially, at identifying gait patterns unique to individual subjects \cite{RN1223, RN746}. The latter study, by Horst \textit{et al.}, reliably demonstrated how LRP could characterise temporal gait patterns, and explained the nuances of particular gait characteristics that distinguished between individual participants in detail \cite{RN746}. An extension of this rationale is that there may be gait patterns that are characteristic of a disease, or diseased sub-population. As such, using the LRP framework, we will attempt to attribute, explain, and interpret the patterns of sensor signal (and therefore the features) that are relevant for distinguishing MS-related gait impairment from healthy ambulation identified using the DCNN models this study (as outlined in figure \ref{fig:tl_overview_schematic}).

\section{Results}
A DCNN model was first trained independently on UCI HAR and WISDM datasets. The information learned from these HAR tasks were then transferred and fine-tuned on the FL dataset for disease recognition (classification) tasks. DCNN models trained exclusively on FL were compared to those fine-tuned from HAR and bench-marked against established feature-based approaches. Finally, DCNN model predictions were decomposed using layer-wise relevance propagation (LRP) in order to interpret the signal characteristics that influenced a prediction for various individual HC and PwMSmod 2MWT segment examples. Table \ref{table:demographics} depicts the population demographics for the FLOODLIGHT PoC dataset. 

\begin{table}[t]
\centering
 \begin{threeparttable}
\caption{Population Demographics of the FLOODLIGHT PoC dataset\tnote{1}. Clinical scores taken as the average per subject over the entire study, where the mean $\pm$ standard deviation across population are reported;  RRMS, Relapsing-remitting MS; PPMS, Primary-progressive MS; SPMS, Secondary-progressive MS; EDSS, Expanded Disability Status Scale; T25FW, the Timed 25-Foot Walk; EDSS (amb.) refers to the ambulation sub-score as part of the EDSS; [s], indicates measurement in seconds; } \label{table:demographics}
\begin{tabular}{m{2cm} M{2.5cm} M{2.5cm} M{2.5cm}}
\hline\midrule
    &
    \parbox{\linewidth}{\centering HC \\(n=24)} & \parbox{\linewidth}{\centering PwMSmild\tnote{a} \\(n=52)} & \parbox{\linewidth}{\centering PwMSmod \tnote{b} \\(n=21)}  \\\midrule
Age                  &  35.6 $\pm$ 8.9 & 39.3 $\pm$ 8.3 & 40.5 $\pm$ 6.9 \\
Sex (M/F)                 &   18/6 & 16/36 & 7/14     \\ 
RRMS/PPMS/SPMS           &    & 52/0/0 & 14/3/4 \\
EDSS                 &   $~$ & 1.7 $\pm$ 0.8 & 4.2 $\pm$ 0.7 \\ 
EDSS (amb.)                &   $~$ & 0.1 $\pm$ 0.3 & 1.9 $\pm$ 1.5 \\ 
T25FW [s]               &   5.0 $\pm$ 0.9 & 5.3 $\pm$ 0.9 & 7.9 $\pm$ 2.2\\\midrule
\end{tabular}
\begin{tablenotes}\footnotesize
\item[1] For more information on the study population we refer the reader to \cite{RN988} and \cite{RN891};
\item[a] PwMS with average EDSS $<3.5$; 
\item[b] PwMS with average EDSS $\geq3.5$; 
    \end{tablenotes}
     \end{threeparttable}
\end{table}

\subsection{Classification Evaluation}
\subsubsection{Evaluation of Activity Recognition}
It was observed that UCI HAR-based activities $(k \in \{walking, stairs, sitting, standing, laying\})$ were well differentiated (Acc: 0.905 $\pm$ 0.018, $\kappa$: 0.880 $\pm$ 0.023, MF\textsubscript{1}: 0.893 $\pm$ 0.025). Much of the confusion between classes occurred between similar ``static'' activities (such as sitting and standing). Distinguishing WISDM-based HAR activities $(k \in \{walking, stairs, sitting, standing, jogging\})$ was less accurate in comparison (Acc: 0.621 $\pm$ 0.037, $\kappa$: 0.525 $\pm$ 0.0047,   MF\textsubscript{1}: 0.622 $\pm$ 0.038), although much of the relative added confusion in WISDM occurred between similar ``dynamic'' activities (such as jogging and walking). Despite this, the prediction of static vs. dynamic activities were distinctly separate for both UCI HAR and WISDM. 

\subsubsection{Evaluation of MS Disease Recognition}
Three separate classification models were constructed for binary tasks (HC vs. PwMSmild, PwMSmild vs. PwMSmod and HC vs. PwMSmod) to allow for direct comparison of the hand-crafted feature-based classification outcomes assessed in \cite{RN988}, as well as a unified multi-class model incorporating all three classes simultaneously. 
The implementation of the baseline SVM model and hand-crafted features have previously been described in \cite{RN988}. Hand-crafted features included various statistical moments of the acceleration epochs and frequency content, as well as energy- and entropy-based properties of the time-frequency signal components though wavelet and empirical mode decomposition.
For further information we refer the reader to \cite{RN988}.\par
Table \ref{table:disease_classification_outcomes} depicts the classification outcomes for all tasks. Bench-marking against a feature based Support Vector Machines (SVM) classifier \cite{RN988}, DCNN (end-to-end) model performance was similar in all tasks. PwMSmod could largely be distinguished from HC and PwMSmild. HAR DCNN models evaluated directly on FL (``direct'') did not distinguish between subject groups. Transfer learning improved disease classification accuracy for all tasks examined relative to feature-based and end-to-end models by upwards of 8\%--15\%, and by 33\% in multi-class tasks, where TL DCNN based on $\mathcal{D}_S$, UCI HAR and WISDM performed similarly for all target classification tasks $\mathcal{T}_T$ (table \ref{table:disease_classification_outcomes}). \par
Further results expanding on this work can be found in the accompanying supplementary material, including the parameters of DCNN models achieving maximal classification performance within table \ref{table:disease_classification_outcomes}.

\begingroup
\setlength{\tabcolsep}{5pt} 
\begin{table}[h!]
\centering
\begin{footnotesize}
 \begin{threeparttable}
 \centering 
 \caption{Comparison of HC vs. PwMS subgroup classification results between various models for each task subset, $\mathcal{T}$. Results are presented as: (1) the posterior overall \textit{subject-wise} outcome for one cross-validation (CV) run as well as (2) the 2MWT \textit{test-wise} median and interquartile range (IQR) across that CV in brackets. The best performing model for each $\mathcal{T}$ are highlighted in \textbf{bold}. Acc: Accuracy; $\kappa$, Cohen's Kappa statistic; MF\textsubscript{1}, Macro-F1 score.}\label{table:disease_classification_outcomes}
\begin{tabular}{l  M{3.5cm} M{3.5cm}  M{3.5cm} M{3.5cm}} \midrule\midrule
 \multirow{2}[3]{*}{\centering $f(\cdot)$} &  Acc. & $\kappa$ &  MF\textsubscript{1}   \\ \cmidrule(l){2-2}  \cmidrule(l){3-3} \cmidrule(l){4-4}
  & \multicolumn{3}{c}{HC vs. PwMSmild} \\  \midrule
  	features + SVM\tnote{1} 									&   0.671 (0.576, 0.544--0.696)	        &  0.212 (0.153, 0.088--0.393) 	        & 0.605 (0.575, 0.527--0.694)  \\
	DCNN (end-to-end)\tnote{2}							        &   0.658 (0.601, 0.517--0.641)	        &  0.226 (0.082, 0.037--0.194)	        & 0.613 (0.541, 0.494--0.588)   \\
	\textbf{DCNN (UCI HAR$\rightarrow$FL)}\tnote{3} 			&  \textbf{0.776 (0.741, 0.688--0.767)}	&  \textbf{0.510 (0.435, 0.346--0.481)}	& \textbf{0.754 (0.716, 0.662--0.737)}   \\
	DCNN (WISDM$\rightarrow$FL)\tnote{3}				        &  0.763 (0.733, 0.698--0.761)	        &  0.486 (0.479, 0.343--0.490)  	    & 0.741 (0.727, 0.667--0.743) 	\\\midrule
   	& \multicolumn{3}{c}{PwMSmild vs. PwMSmod} \\\midrule
	features + SVM\tnote{1} 									&  0.849 (0.783, 0.706--0.858) 	        & 0.627 (0.566, 0.412--0.708)  	        & 0.813 (0.778, 0.692--0.853)  \\
	DCNN (end-to-end)\tnote{2} 								    &  0.822 (0.682, 0.617--0.763)  	    & 0.583 (0.356, 0.166--0.444)	        & 0.791 (0.675, 0.562--0.721)   	\\
	DCNN (UCI HAR$\rightarrow$ FL)\tnote{3}				        &  0.904 (0.849, 0.839--0.873)          & 0.776 (0.675, 0.650--0.707)  	        & 0.888 (0.837, 0.823--0.852)    \\
	\textbf{DCNN (WISDM$\rightarrow$FL)}\tnote{3} 			    &  \textbf{0.918 (0.869, 0.833--0.935)} & \textbf{0.810 (0.690, 0.630--0.844)}  &  \textbf{0.905 (0.845, 0.812--0.922)}  \\\midrule
   & \multicolumn{3}{c}{HC vs. PwMSmod} \\\midrule
	features + SVM \tnote{1}								    &  0.800 (0.773, 0.737--0.881) 	        &  	0.595 (0.546, 0.474--0.763)  	    &  0.796 (0.772, 0.737--0.881) 	\\
	DCNN (end-to-end)\tnote{2} 								    &  0.822 (0.734, 0.663--0.831)          &   0.641 (0.462, 0.292--0.657)         &  0.820 (0.730, 0.618--0.828) 	\\
	DCNN (UCI HAR$\rightarrow$FL)\tnote{3}				        &  0.889 (0.873, 0.730--0.929)          &    0.777 (0.743, 0.446--0.847)	    &  0.889 (0.870, 0.723--0.924)     \\
	\textbf{DCNN (WISDM$\rightarrow$FL)}\tnote{3} 		        &  \textbf{0.911 (0.886, 0.766--0.911)} & \textbf{0.821 (0.772, 0.520--0.820)}  &  \textbf{0.911 (0.886, 0.760--0.910)} 	\\\midrule
		   & \multicolumn{3}{c}{HC vs. PwMSmild vs. PwMSmod} \\\midrule
	features + SVM\tnote{1}									    &  0.629 (0.551, 0.510--0.577) 	        &  	0.368 (0.093, 0.020--0.103)  	    &  0.580 (0.510, 0.495--0.540) 	\\
	DCNN (end-to-end)\tnote{2} 								    &  0.608 (0.503, 0.488--0.516) 	        &  	0.274 (0.106, 0.081--0.130)  	    &  0.523 (0.446, 0.402--0.483) 	\\
	\textbf{DCNN (UCI HAR$\rightarrow$FL)}\tnote{3}	            &  \textbf{0.814 (0.703, 0.700--0.744)} &  \textbf{0.673 (0.331, 0.325--0.423)} &  \textbf{0.796 (0.672, 0.664--0.720)}      \\
	DCNN (WISDM$\rightarrow$FL)\tnote{3}				        &  0.763 (0.690, 0.677--0.737)          &   0.571 (0.303, 0.274--0.407) 	    &  0.725 (0.671, 0.644--0.699)  	\\\midrule
\end{tabular}
\begin{tablenotes}\footnotesize
\item[1] ``features + SVM'' refers to classification performed using features and a SVM with the pipeline described in \cite{RN988}; 
\item[2] ``end-to-end'', refers to a model trained and validated end-to-end exclusively on $\mathcal{D}_T$ data;
\item[3] ``$\rightarrow$''  denotes the source HAR dataset $\mathcal{D}_S$ used and transferred to FL $\mathcal{D}_S$ and $\mathcal{T}_S$.\\ 
\item [] See figure \ref{fig:tl_schematic} for a more detailed description of the TL approach used in this study. 
\end{tablenotes}
\end{threeparttable}
\end{footnotesize}
\end{table}
\endgroup

\subsection{Interpreting MS Remote Sensor Data}\label{sec:LRP_results}
The results described in this section aim to interpret smartphone sensor data recorded from FL through attribution techniques. Trained models were decoded using LRP, where we propose that this framework allows us to understand (at least to some extent) the classification decision in individual out-of-sample 2MWT epochs. \par
Holistic interpretation of the disease-classification outcomes with respect to the inertial sensor data can be greatly augmented from the integration of: (1) visualising the raw data, (2) its time-frequency representation using the (discretised) continuous wavelet transform (CWT) and (3) LRP attribution techniques. The CWT is a method used to measure the similarity between a signal and an analysing function (in this case the Morlet wavelet) which can provide a precise time-frequency representation of a signal \cite{RN988, RN701}. For more information we refer the reader to the analysis performed in \cite{RN988} \par
\begin{figure*}[t!]
    \centering
     \includegraphics[width=1\linewidth]{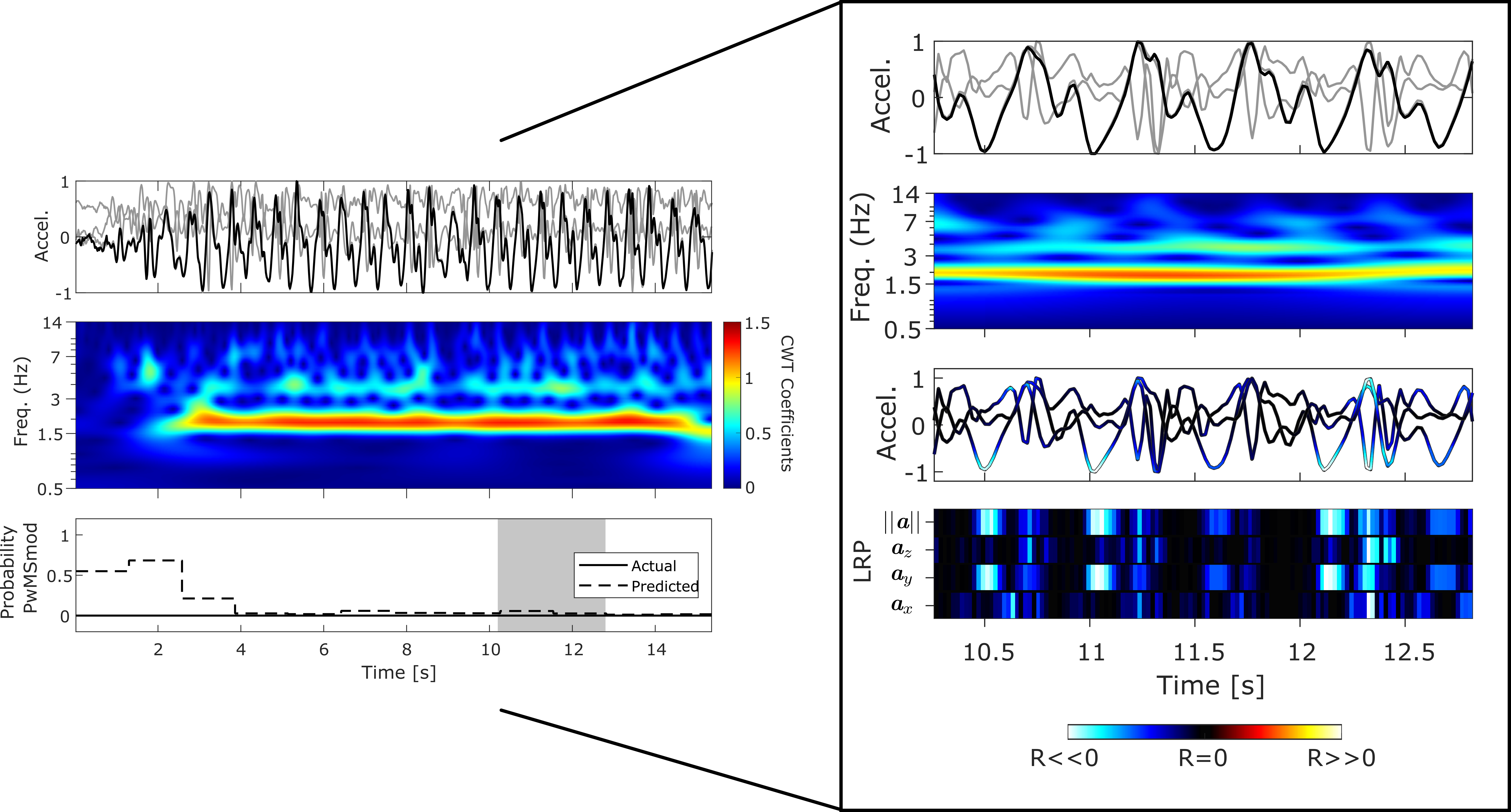}
    \caption{\footnotesize \textbf{HC epoch: Panel plot illustrating example performance of a typical HC subject (true negative) which can be visually~interpreted using LRP decomposition and CWT frequency analysis.} (HC, T25FW: 4.8 $\pm$ 0.35) \\
    The top row represents a 3-axis accelerometer trace captured from a smartphone over 15.4 seconds, which corresponds to 12 epochs of length 128 samples (or 2.56 [s]) with a 50\% overlap. The magnitude $(\|\textit{\textbf{a}}\|$) of the 3-axis signal is highlighted in bold. The second row depicts the top view of the CWT scalogram of $\|\textit{\textbf{a}}\|$, which is the absolute value of the CWT as a function of time and frequency. The final row depicts the output disease classification probabilities ($\mathcal{T}_T$). The shaded grey area represents an example epoch (n=128 samples, or 2.56 [s]) within the acceleration trace, which is examined in the larger subplot through the decomposition of DCNN input relevance values ($R_i$) using Layer-Wise Relevance Propagation (LRP). Red and hot colours identify input segments denoting positive relevance ($R_i>0$) indicating $f(\bm{x})>0$ (i.e. MS). Blue and cold hues are negative relevance values ($R_i<0$) indicating $f(\bm{x})<0$ (i.e. HC), while black represents ($R_i\approx0$) inputs which have little or no influence to the DCNNs decision.  LRP values are overlaid upon the accelerometer signal, where the bottom panel represents the LRP activations per input (i.e. $\textit{\textbf{a}}_x, \textit{\textbf{a}}_y, \textit{\textbf{a}}_z, \|\textit{\textbf{a}}\|$). }\label{fig:lrp:HC_TN}
  \end{figure*}

\begin{figure*}[t!]
    \centering
    \includegraphics[width=1\linewidth]{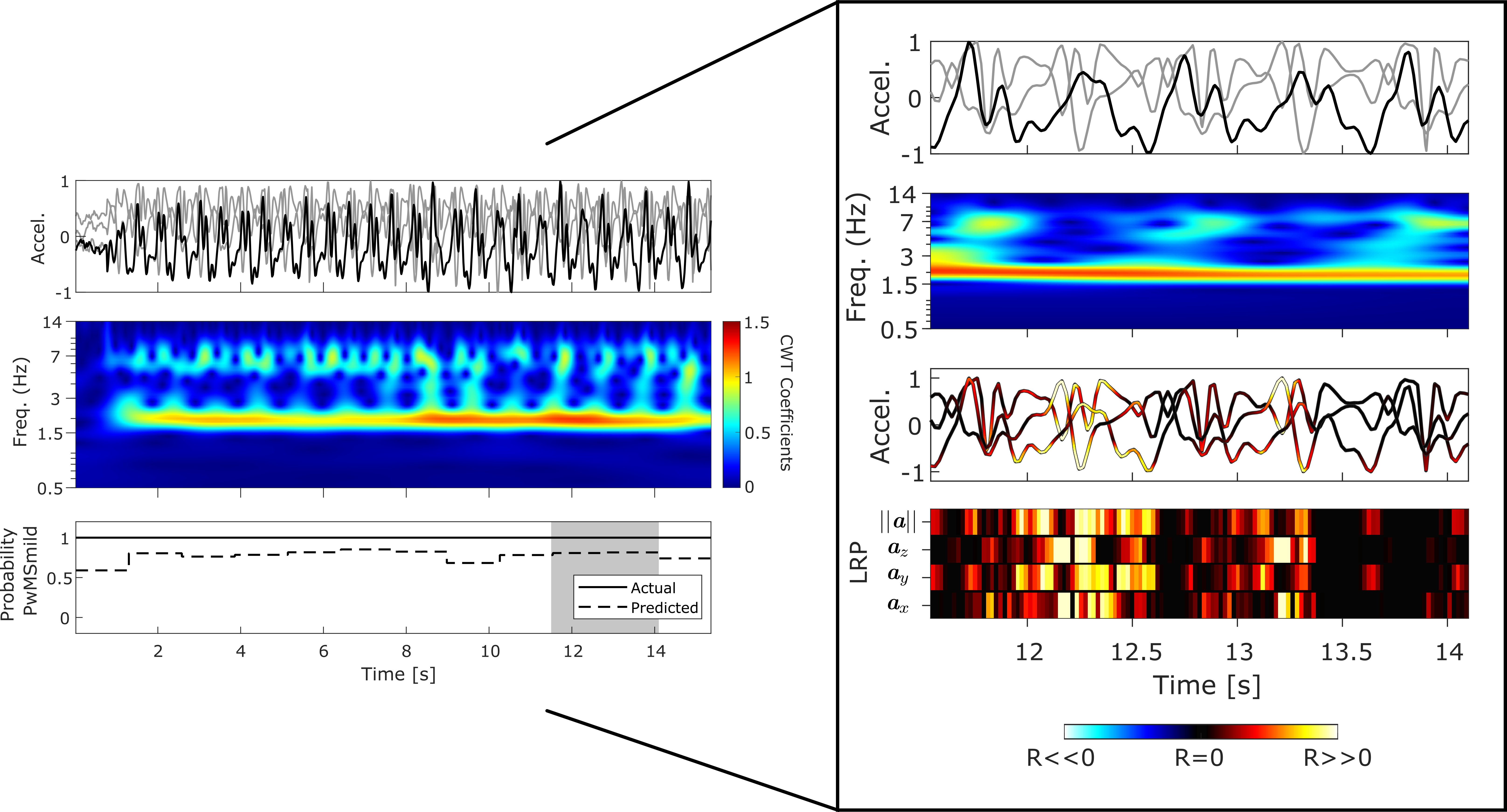}
    \caption{\footnotesize 
    \textbf{PwMSmild epoch: Panel plot illustrating example performance from a section of correctly classified PwMSmild subject's 2MWT (true positive) which can be visually~interpreted using LRP decomposition and CWT frequency analysis.} (PwMSmild, EDSS 3.25 $\pm$ 0.35; T25FW: 5.5 $\pm$ 0.53 [s]) For further information regarding the interpretation of this example, we refer the reader to figure \ref{fig:lrp:HC_TN}. 
 }\label{fig:lrp:PwMSmild_TP}
\end{figure*}
Relevance propagation through LRP decomposed the output $f(\bm{x})$ of a learned function $f$, given an input $\bm{x}$, attributing relevance values $R_i$ to individual input samples $x_i$. In this case, $x_i$ were represented by discrete sensor samples within an testing epoch and therefore $R_i$ was directly embedded in the time domain. The contribution of LRP could also be quantified across the input channels (in this case the sensor axis). 
\par Figures \ref{fig:lrp:HC_TN}--\ref{fig:lrp:PwMSmod_TP} compared the example patterns and characteristics captured from depicting the raw sensor signal, augmented with LRP-CWT analysis for representative examples of correctly classified HC, PwMSmild and PwMSmod subjects respectively.
Figure \ref{fig:lrp:HC_TN} first illustrated the performance of a correctly classified HC subject's 2MWT segment, supplemented by the raw sensor data, its CWT representation and the final disease model's probabilistic output $p(\bm{x})$. In this example, gait signal was apparent from the raw sensor data and supported by strong gait domain energy, $E_s$, within the CWT representation. This collection of epochs were predicted as predominantly ``walking'' by a HAR model and corresponded to a confident HC classification with high probability. Focusing on an overlapped epoch example from 10.3--12.8 [s], gait was clearly visible in the time-frequency domain (i.e. large CWT coefficients around 1.5 Hz) and reflected clear steps, as depicted by the magnitude of acceleration. LRP attributed high relevance scores $R_i$ to these steps in the vertical $\textit{\textbf{a}}_y$ and orientation invariant signal $||\textit{\textbf{a}}||$ (i.e. represented by channel 2 and 4). 
\par Figure \ref{fig:lrp:PwMSmild_TP} depicted an example 2MWT from a representative correctly classified PwMSmild subject. Similarly to the HC example, gait signal was visible in this example, (i.e. CWT $E_s$, HAR $f(\bm{x})$ and clear steps in $||\textit{\textbf{a}}||$) (Note: HAR posterior probabilities also indicated ``walking''). Relevance propagation for an example epoch during 11.5--14 [s] indicated that step occurrences attributed to the ``mild'' DCNN posterior output. Time-frequency gait signal visualisation through CWT analysis also revealed harmonics occurring at higher frequencies than the gait domain ($>$3.5 Hz). 
\par Figure \ref{fig:lrp:PwMSmod_TP}, in contrast, represented a panel plot illustrating example performance of a typical correctly classified PwMSmod subject's 2MWT segment. In this example, a concentration of higher frequency $E_s$ disturbances temporally coincided with each step event. These gait-related perturbations were examined in the zoomed sub figure for an example epoch during 3.9--6.4 [s], as highlighted by the shaded shaded grey area during the longer gait example. Relevance decomposition of this epoch attributed all LRP-based relevance $R_i$ to each step and associated high-frequency disturbance (i.e. events influencing $f(\bm{x})>0$ output as PwMSmod).\par
\par Finally, average gait epochs for HC, PwMSmild and PwMSmod groups were created using Dynamic Time Warping Barycenter Averaging \cite{RN1203}. Visualisation of each representative epoch using the LRP-CWT framework was depicted in figure \ref{fig:average_epoch}. The DCNN posterior probabilities for each representative epoch were strongly  predictive of the true class (HC, Pr. 0.89; PwMSmild, Pr. 0.91; and PwMSmod, Pr. 0.90). 

\section{Discussion}
\subsection{Learning a Representation of MS Ambulatory Function}
Deep networks may learn a better representation of gait function collected from smartphone-based inertial sensors, than those of traditionally hand-crafted features. This study leveraged DCNNs to extract unconstrained features on raw smartphone accelerometery data captured from remotely performed 2MWTs by HC and PwMS subjects. Rather than relying on hand-crafted features, which are constrained transformations and approximations of the original signal, DCNNs offer a data-driven approach to characterise ambulatory related features directly from the raw sensor data. Remote health data is often sparse and infrequently sampled \cite{RN988}; low study participant numbers and heterogeneous data can make it difficult to build reliable and robust models. To help overcome this we have demonstrated how we can learn common gait-related characteristics from open-source datasets first, then fine-tune these representations to learn disease-specific ambulatory traits using transfer learning. \par
In this work, higher-level DCNN features learned on open-source HAR datasets were transferred to the FL domain. A DCNN model was first trained as a HAR classification task on two independent open-source HAR datasets. In accordance with other studies \cite{RN996}, excellent discrimination of HAR activities was achieved using deep networks in the UCI HAR database. Prediction of WISDM-based HAR activities were less accurate in comparison, although much of the relative added confusion in WISDM occurred between similar ``movement'' activities (such as jogging and walking). Importantly, applying a HAR-trained model directly to FL did not identify HC and PwMS subgroups.\par
The results presented in this study demonstrated that DCNN models can be applied to raw smartphone-based inertial sensor data to successfully distinguish sub-groups of PwMS from HC subjects. The performance of DCNN models applied directly to FL (end-to-end) was similar to that of the feature-based approaches.
However, models that were trained end-to-end using only FL data were highly susceptible to over-fitting and struggled to generalise compared to models which had been fine-tuned from a trained HAR network. This paradigm was particularly evident in the poor end-to-end model performance for the more difficult multi-class $\mathcal{T}_T$ task.  As a result, transfer learning improved model robustness and generalisability towards the recognition of subgroups of PwMS from HCs, compared to DCNN models trained end-to-end on FL, as well as the feature based methods (by between 8\%--15\% in binary tasks and up to 33\% in the multi-class task). Larger performance gains in the latter were particularly evident as improved recognition of PwMSmild from HC.  \par
\begin{figure*}[t!]
    \centering
    \includegraphics[width=1\linewidth]{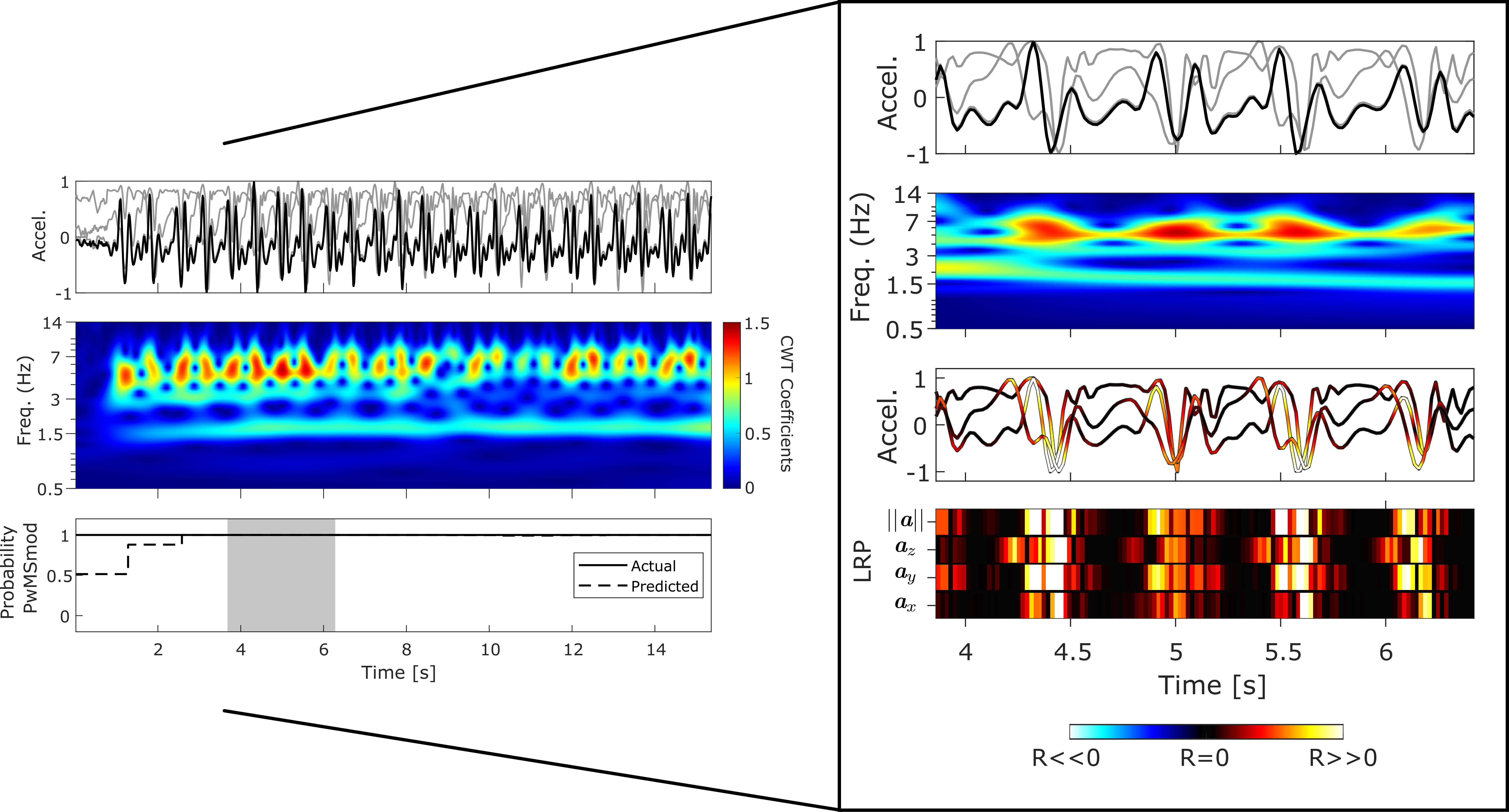}
    \caption{\footnotesize 
    \textbf{PwMSmod epoch: Panel plot illustrating example performance from a section of a correctly classified PwMSmod subject's 2MWT (true positive) which can be visually~interpreted using LRP decomposition and CWT frequency analysis.} (PwMSmod, EDSS: 3.8 $\pm$ 2.9; T25FW: 6.9 $\pm$ 0.5 [s]) For further information regarding the interpretation of this example, we refer the reader to figure \ref{fig:lrp:HC_TN}.
 }\label{fig:lrp:PwMSmod_TP}
\end{figure*}
The improved DCNN model performance through TL could be attributed to a number of rationale. For instance, there is added benefit of training on a larger and more diverse set of data, as well as the regularisation properties TL induces (freezing layers mitigates against over-fitting). More interestingly, both UCI HAR (n=30) and WISDM (n=51) specifically comprised of young healthy individuals. In FL there were only n=24 healthy participants (only 16 of which contributed more than 10 unique running belt 2MWT tests). As such, initially training on more healthy examples in particular may have allowed the DCNN to initialise a more accurate representation of ``healthy'' walking from inertial sensor data. It is noteworthy that models transferred based on $\mathcal{D}_S$ UCI HAR versus WISDM performed similarly for all target classification tasks $\mathcal{T}_T$. Transferring from WISDM tended to perform marginally better at distinguishing PwMSmod however, whereas PwMSmild were slightly better identified from HC when transferring from UCI HAR. Perhaps the activity patterns within each dataset also uniquely aided each task. For example, WISDM explicitly learned a unique ``jogging'' class, which could allow the better representation learning of faster versus slower gait. Moderately disabled PwMS in particular are known to have relatively slower cadence \cite{RN924}. Other characteristics must also be considered, such as affixing of the phone to the waist in UCI HAR (similar to the FL running belt) versus the pocket in WISDM. Regardless, more work is certainly needed to uncover the performance gain and understand explicit reasons for the improvement of TL models applied to remote sensor data. Particularly, further studies will be needed to fully define the attributes of a source domain $\mathcal{D}_S$ and task $\mathcal{T}_S$ which are relevant for the target domain $\mathcal{D}_T$ and tasks $\mathcal{T}_T$, or to determine the optimal $\mathcal{D}_S$ (or combination thereof) among multiple $\mathcal{D}_S$ candidates.\par 
The DCNN architecture investigated in this study was relatively simple compared to other frameworks \cite{RN763, RN989, RN737, RN735}, future work will also aim to investigate the use recurrent layers (such as in LSTMs), which have proven beneficial to characterise the temporal nature of gait recorded within the sensor signal \cite{RN735, RN737}.\par

\subsection{Interpreting Smartphone-based Remote Sensor Models}
Recently, breakthroughs in visualising neural networks have paved the way for explanations in deep and complex models, for example heatmaps of ``relevant'' parts of an input can be built by decomposing the internal neural networks using layer-wise relevance propagation \cite{RN835, RN948}. Visual interpretation of the factors which influence a model's prediction may enable a deeper understanding of how healthy and disease-influenced characteristics are captured from remote smartphone-based inertial sensor data. This work aims to 
establish a framework to further understand the patterns of healthy and MS disease through multiple viewpoints: visualising the raw inertial sensor signal, its analogous time-frequency CWT representation, as well as augmenting this picture using layer-wise relevance propagation. Attribution through LRP has already been successfully applied to visualise gait patterns that are predictive of an individual which were acquired from lab-based ground reaction force plates and infrared camera-based full-body joint angles \cite{RN746}. \par
The patterns of healthy gait were first visually inspected in an example HC 2MWT (figure \ref{fig:lrp:HC_TN}) through the LRP-CWT framework. Comparing these healthy gait templates to PwMS examples offered a visual interpretation between the differences in the signal characteristics each classifier recognised as important for a prediction. For instance, it was found that walking in healthy predicted gait in FL was typically characterised by distinct steps, consistent cadence, and strong gait power ($E_s$) in the 1.5--3 Hz range. Attribution using LRP highlighted step inflections, especially in the vertical $\textit{\textbf{a}}_y$ and magnitude of acceleration signals $||\textit{\textbf{a}}||$, as important predictors for HC ambulation.  
Misclassified PwMSmild and PwMSmod examples as HC depicted in this work (see supplementary material) also tended to visually resemble that of the HC (figure \ref{fig:lrp:HC_TN}), for instance LRP tended to attribute relevance to the $\textit{\textbf{a}}_y$ and $||\textit{\textbf{a}}||$ channels during clear step inflections, much like to that of the actual HC example. \par
\begin{figure*}[t]
 \centering
 \begin{subfigure}[c]{0.32\linewidth}
 \includegraphics[width=0.95\linewidth]{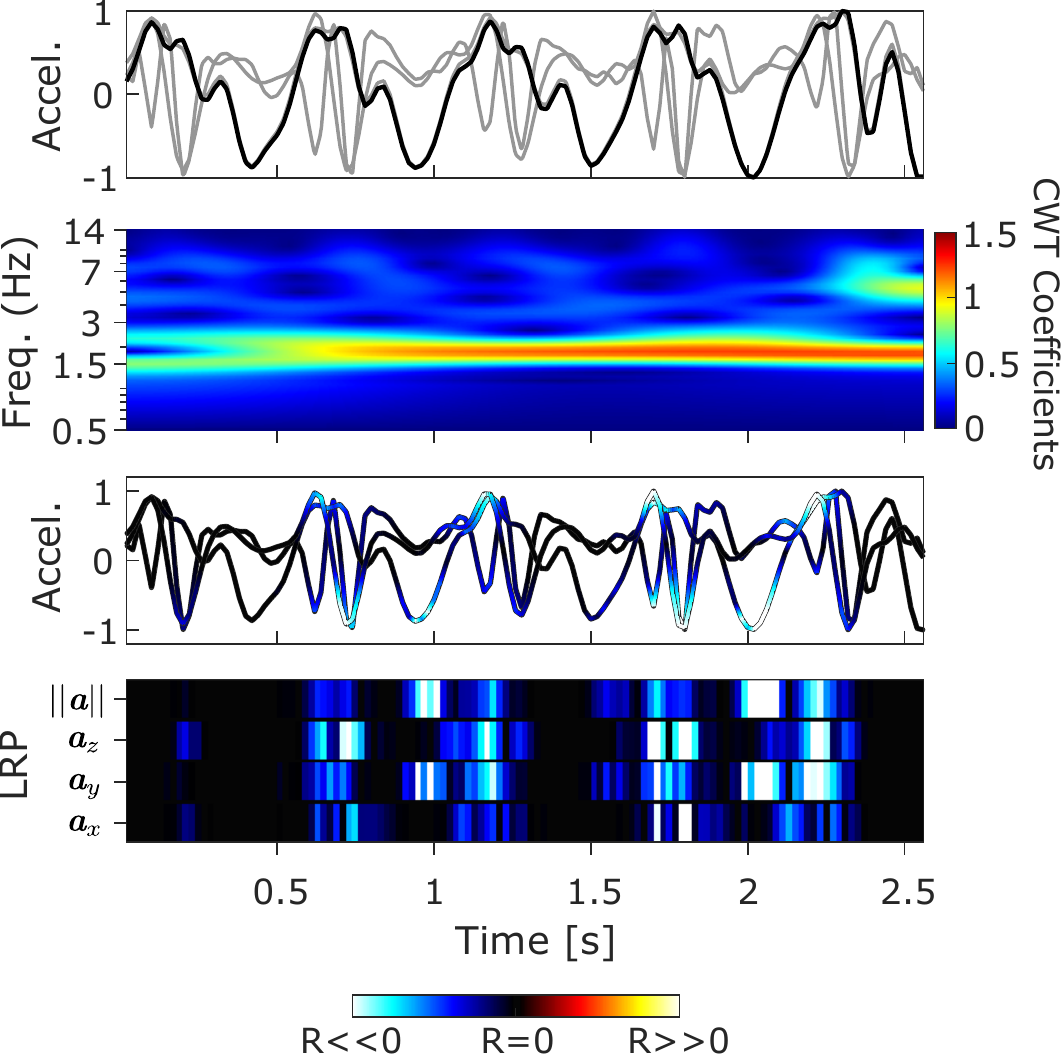}
 \caption{HC (Pr: 0.89)}\label{fig:average_epoch_HC}
\end{subfigure}
 \begin{subfigure}[c]{0.32\linewidth}
 \includegraphics[width=0.95\linewidth]{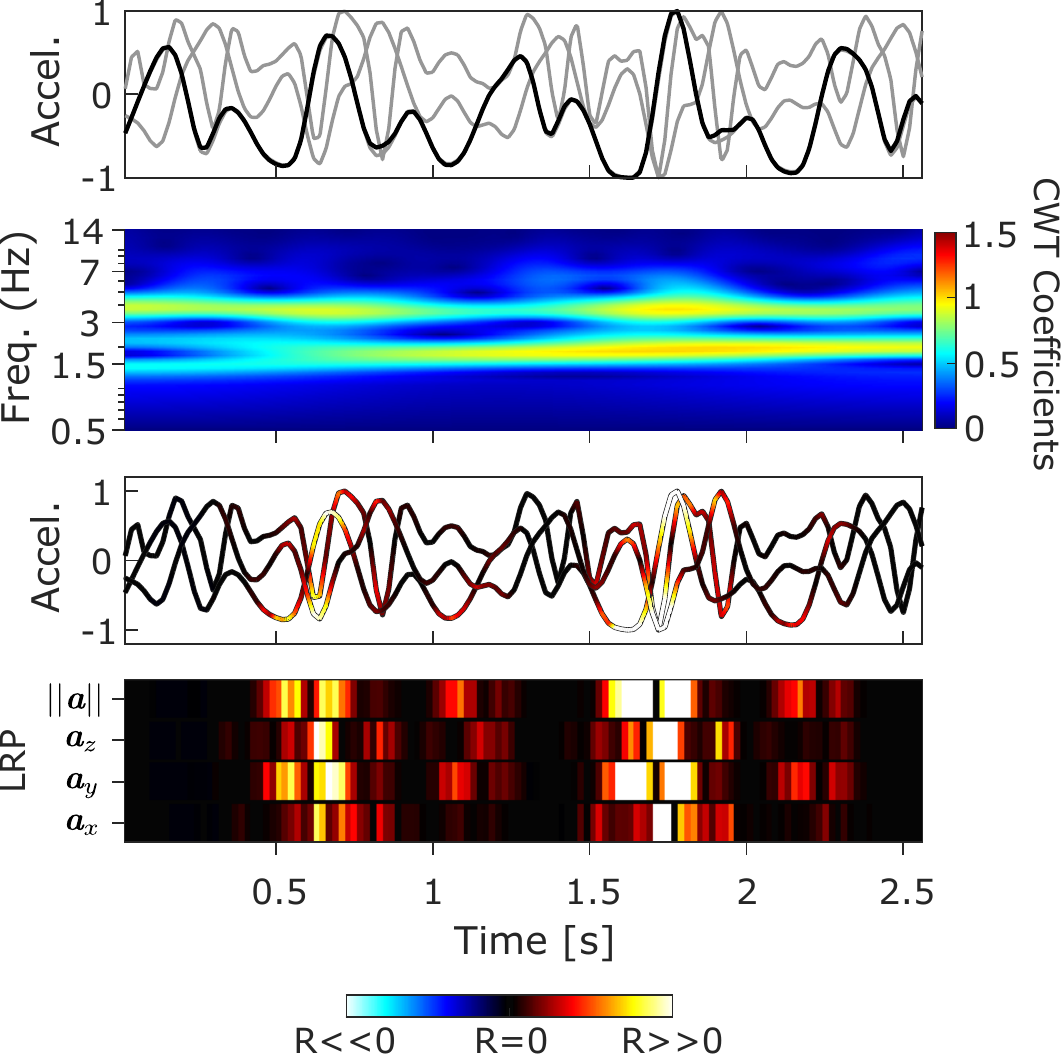}
  \caption{PwMSmild (Pr: 0.91)}\label{fig:average_epoch_PwMSmild}
\end{subfigure}
 \begin{subfigure}[c]{0.32\linewidth}
 \includegraphics[width=0.95\linewidth]{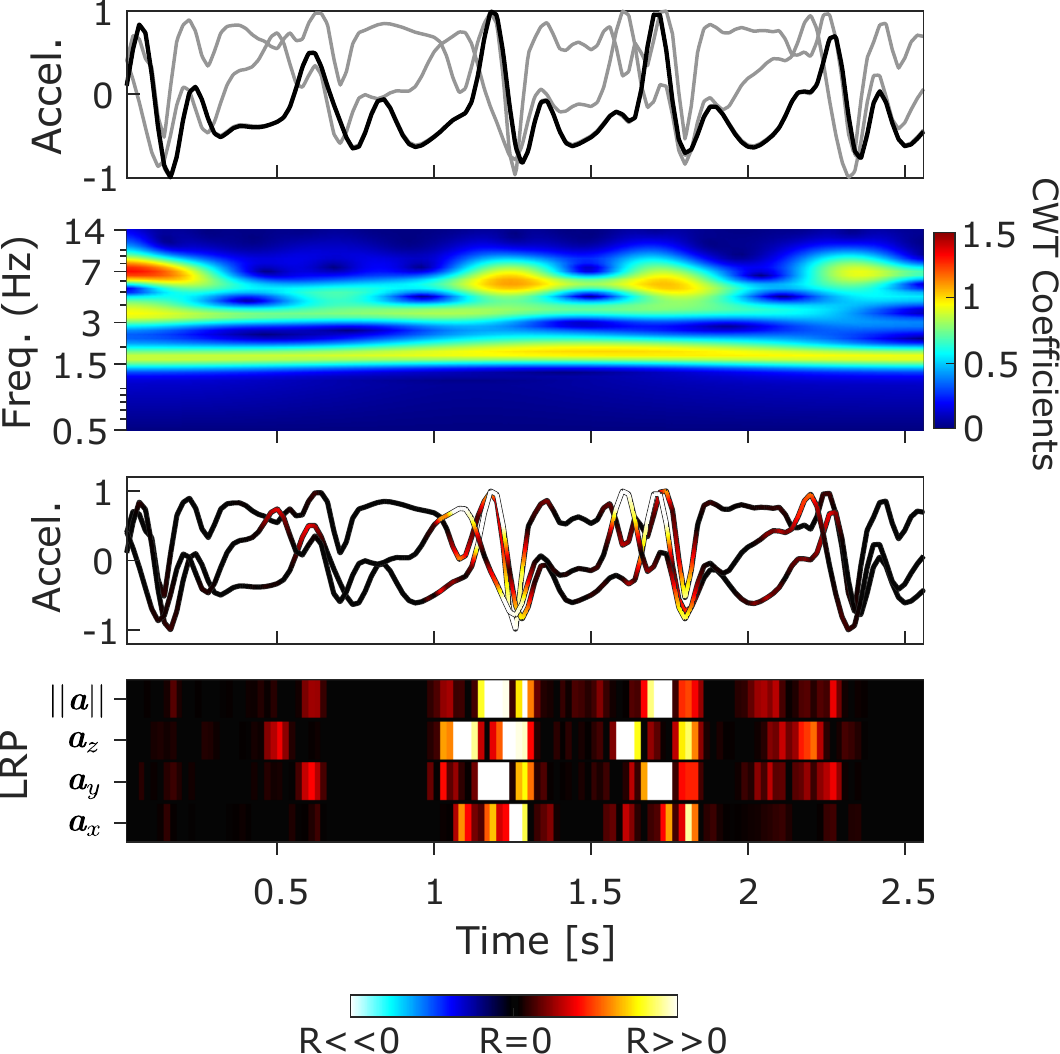}
  \caption{PwMSmod (Pr: 0.90)}\label{fig:average_epoch_PwMSmod}
\end{subfigure}
\caption{\textbf{Visualisation of the average gait signal through the LRP-CWT analysis framework.} Representative average epochs (n=128 samples, 2.56 [s]) were first created using Dynamic Time Warping Barycenter Averaging (DBA) independently for sets of correctly classified epochs from HC, PwMSmild and PwMSmod subject groups. (Pr; represents the DCNN posterior probability for that class). }\label{fig:average_epoch}
\end{figure*}
The morphology of the sensor data in PwMSmod examples were visually different to correctly classified HC and PwMSmild. In the case of the PwMSmod gait epochs (figure \ref{fig:lrp:PwMSmod_TP}) and the false positives (see supplementary material), these examples exhibited distinct higher frequency ``pertubations'' in the presence of gait, where further LRP decomposition attributed those disturbances as temporally important for the prediction of PwMSmod in each case.  Interestingly, these higher frequency $E_s$ disturbances temporally coincided with each step event from the raw sensor signal. Misclassified examples of HC and PwMSmild as PwMSmod tended to reflect similar properties to the correctly classified example from figure \ref{fig:lrp:PwMSmod_TP}, such as as lower $E_s$ gait domain energy (see \cite{RN988}), evidence of higher frequency perturbations and a visually less well defined step morphology within the raw sensor signal. As corroboration to the factors influencing these misclassifications, LRP clearly attributed positive relevance (i.e. PwMSmod predictions) to time points of the signal corresponding to higher frequency signal-based activity.
\par Generating representative correctly classified gait epochs using DTW Barycenter Averaging painted a macro-picture of the average gait patterns for HC, PwMSmild and PwMSmod groups. Visualising these representative epochs using the LRP-CWT framework (figure \ref{fig:average_epoch}), displayed confirmatory patterns observed in previous independently classified 2MWT examples (figures \ref{fig:lrp:HC_TN}--\ref{fig:lrp:PwMSmod_TP}). Importantly, the raw accelerometer signals collected from healthy controls and the DBA-generated average cycles were highly comparable to established characteristics that are representative of healthy walking \cite{RN171, RN1215}. For example, a DBA representative HC epoch was visually observed to have clear step patterns with discernible initial and final feet contact points, and stronger $E_s$ in the gait domain (0.5--3 Hz). In comparison, milder and moderate MS-predicted average epochs tended to have reduced (gait) signal-to-noise $E_s$ and the presence of higher-frequency perturbations, as described previously. 
\par The uncovering of the inertial sensor characteristics that distinguished MS-related disease from healthy ambulation, through LRP attribution, enables for clinical interpretations. For example, higher relevance values coinciding with distinct step inflections and higher power ($E_s$) in the upper-end of the gait domain could represent cadence-based factors associated with faster walking, which are established characteristics that may differentiate healthy versus PwMS ambulation \cite{RN1159, RN924, RN289}. Moderately disabled PwMS in particular are known to have relatively slower cadence \cite{RN924}. The attribution of gait disturbances suggesting MS-related impairment, could be associated with other accepted indicators, such as gait variability, which have shown to stratify PwMS from HC \cite{RN293,RN710,RN728}. \par
More interestingly, hand-crafted features that captured similar characteristics to those visually observed through the CWT-LRP framework appeared before as top features within our previous study \cite{RN988}. Features such as wavelet entropy and energy, capturing predictability and energy in the faster gait domain (1.5--3.3 Hz), or (gait) signal-to-noise related measures separated the same healthy and PwMS participants. 
\par The similarity between hand-crafted features, visual examples, and LRP-explained DCNN features clinically corroborated an interpretation of the factors which may be sensitive to MS-related gait impairment. The hand-crafted features introduced previously in \cite{RN988} focused on using established signal-based metrics as surrogates to represent aspects of PwMS gait function. As such, these surrogate features were not engineered to specifically capture complex biomechanical processes in PwMS gait. Data-driven measures may therefore have been more comprehensive, sensitive, or specific to capture the same representation of MS-indicative characteristics, than the approximations from constrained, hand-crafted features.

\subsection{Limitations and Concluding Remarks}
There are a number of limitations which should be discussed with respect to this study. 
Firstly, while deep networks exhibit unrivalled potential in many healthcare applications, such as in this setting to characterise ambulatory and physical activity patterns from wearable accelerometery, the ramifications of applying these models to healthcare data should also be considered. Often observational clinical studies are small and initially collecting vast amounts of data on a larger number of participants can be both unfeasible and costly. 
Although the TL-framework introduced in this work helps overcome some of the difficulties encountered when attempting to build deep networks in the presence of heterogeneity and low subject numbers, the fine-tuning and evaluation of DCNN performance could still be predisposed by the limitations of the data. For instance, the relatively small number of participants (n$<$100) with multiple repeated measurements, the differences in the number of unique tests contributed per subject, or even demographic biases, such as the male-to-female ratio mismatches between HC and PwMS, the inclusion of various different MS phenotypes, or that the mild versus moderate sub-groups were bluntly created based on clinician-subjective EDSS scores, are all factors that should all be considered when evaluating model performance. Learning more accurate global models were therefore biased by the diversity, representation, and size of the data available.
In reality, the NCT02952911 FLOODLIGHT PoC study was only intended as a small proof-of-concept investigation to assess the feasibility to remotely monitor PwMS subjects, yet has provided many meaningful insights which can be implemented in future studies. Follow-up trials with larger, more diverse cohorts are already being undertaken, such as FLOODLIGHT OPEN, a crowd sourced dataset where the general public can contribute their own data \cite{van2019floodlight}. \par 
Despite the clinical utility LRP could hold in visualising and interpreting neural network decisions, heatmap interpretations were only qualitatively evaluated based on visual assessment, albeit motivated by a clinical hypothesis. For instance, LRP relevance values coinciding with distinct step inflections and higher power ($E_s$) in the upper-end of the gait domain could represent cadence-based factors associated with faster walking, which are established characteristics that may differentiate healthy versus PwMS ambulation \cite{RN1159, RN924}. Other objective methods to evaluate heatmap representations have been proposed which involve perturbing the model's inputs \cite{RN967}. Verifying that LRP has attributed meaningful relevance is inherently difficult however due to the remote nature of the 2MWTs performed by participants in this study. 
Further studies should aim to evaluate HC and PwMS gait function in more controlled settings, such as under visual observation or using in-clinic gait measurement systems, which will allow the underlying attributions of LRP to be further evaluated.
\par
More comprehensive analysis should also aim to compare various other attribution techniques, especially similar attribution methods, to evaluate smartphone-based remote sensor models. This future work could be used to further verify the predictive patterns uncovered with one method (e.g. the confirmatory hypothesis of another attribution method also picking up on the same pattern as LRP highlighted). \par
As initial steps, this study focused on establishing clear and concise interpretations of smartphone-based inertial sensor models to characterise patterns of gait and disease-influenced ambulation by first focusing on only the positive contributions towards class predictions (i.e. LRP $\alpha_1\beta_0$-rules). With this baseline, further work (and especially in more controlled settings) should aim to apply LRP for more complex tasks to develop more full-bodied explanations, such as visualising contributions which contradict the prediction of a class (e.g. using LRP $\alpha_2\beta_1$-rules). 

\par In conclusion, the work presented here aimed to explore the ability of deep networks to detect impairment in PwMS from remote smartphone inertial sensor data. Transfer learning may present a useful technique to circumvent common problems associated with remotely generated health data, such as low subject numbers and heterogeneous data. TL DCNN models appeared to learn a better representation of gait function compared to feature based approaches characterising HCs and subgroups of PwMS. Further work is needed however to to understand the underlying feature structure, along with the most applicable source datasets and methods to extract the most appropriate information available. Incorporating expert clinical knowledge through better visual interpretation techniques could greatly develop clinicians’ fundamental understanding of how disease-related ambulatory activity can be captured by wearable inertial sensor data. This work proposed the use of LRP heatmaps to interpret a deep network's decisions by attributing relevance scores to the inertial sensor data and augmenting this assessment with time-frequency visual representations. This improved domain knowledge could be used to reverse engineer features, develop more robust models or to help refine more sensitive and specific measurements. This study, with on-going future work, therefore further demonstrates the clinical utility of objective, interpretable, out-of-clinic assessments for monitoring PwMS. 

\section{Methods}\label{sec:Methods}
\subsection{Data}
The FLOODLIGHT (FL) proof-of-concept (PoC) study (NCT02952911) was a trial to assess the feasibility of remote patient monitoring using smartphone (and smartwatch) devices in PwMS and HC \cite{RN891}. A total of 97 participants (24 HC subjects; 52 mildly disabled, PwMSmild, EDSS [0-3]; 21 moderately disabled PwMSmod, EDSS [3.5-5.5]) contributed data which was recorded from a 2MWT performed out-of-clinic \cite{RN988}. 
Subjects were requested to perform a 2MWT daily over a 24-week period, and were clinically assessed baseline, week 12 and week 24. For further information on the FL dataset and population demographics we direct the reader to \cite{RN891} and specifically to our previous work \cite{RN988}, from which this study expands upon.
\begin{figure}[b!]
	\centering
  \includegraphics[width=0.75\linewidth]{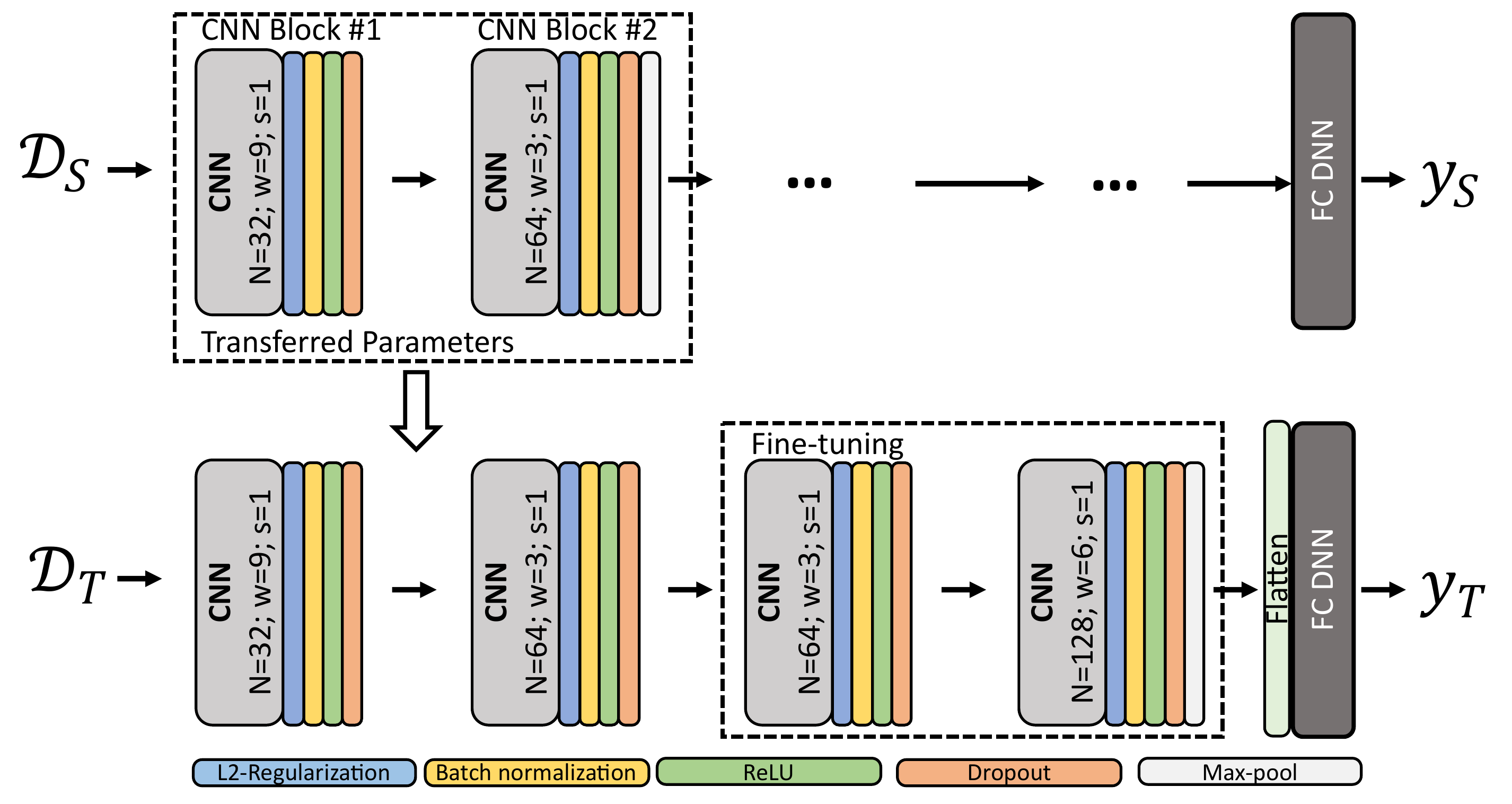} 
\caption{\footnotesize \textbf{Schematic of deep transfer learning approach}. $\mathcal{D}_S$ refers to input data from a source domain, in this case a HAR dataset, to learn a task $\mathcal{T}_S$, which is represented by the label space $\mathcal{Y}_S$ (the HAR activity classes).  $\mathcal{D}_T$ refers to the target domain, in this case the FLOODLIGHT data, where $\mathcal{Y}_T$ are the disease classification outputs of HC, PwMSmild or PwMSmod for target task $\mathcal{T}_T$. During transfer learning, a model's parameters and learned weights, $f(\cdot)$ of $\mathcal{D}_S$, are then used to initialise and train a model on target domain $\mathcal{D}_T$ and task $\mathcal{T}_T$. Transfer learning is then performed by transferring the source model's layers (where these weights and parameters are ``frozen'') to subsequently re-train a new model (i.e. fine-tune) using $\mathcal{D}_T$ data for the new target task, $\mathcal{T}_T$. Downstream layers in the network are fine-tuned towards this new target task decision $\mathcal{Y}_S$. }\label{fig:tl_schematic} 
\end{figure} 

\subsection{Deep Transfer Learning for Time Series classification}\label{sec:DL_for_timeseries}
\subsubsection{Model Construction}
In this time series classification problem, raw smartphone sensor data recorded during remote 2MWTs were partitioned into epoch sequences and DCNNs were used to classify each given sensor epoch, $\mathbf{X}_n$, as having been performed by a HC, PwMSmild or PwMSmod participant; where $\mathbf{X}_n =\left(\textit{\textbf{a}}_x, \textit{\textbf{a}}_y, \textit{\textbf{a}}_z, \|\textit{\textbf{a}}\|\right)^\top$, \textit{\textbf{a}} are acceleration vectors for the $x$-, ${y}$- and $z$-axis coordinates, containing samples $\textit{\textbf{a}}=\left(x_1, x_2, ..., x_T\right)$ and $\|\textit{\textbf{a}}\|$ refers to original orientation invariant signal magnitude. \par
Three separate classification models were constructed for binary tasks (HC vs. PwMSmild, PwMSmild vs. PwMSmod and HC vs. PwMSmod) to allow for direct comparison of the hand-crafted feature-based classification outcomes in \cite{RN988}, as well as a unified multi-class model incorporating all three classes simultaneously. The population subset explored for this study are the same as reported previously in \cite{RN988}. The accompanying supplementary material further details the DCNN model architecture, parameterisation, and evaluation. 
\par A model architecture was first trained on source domain $\mathcal{D}_S$ and task $\mathcal{T}_S$, in this instance a HAR classification task on the UCI HAR or WISDM dataset (see supplementary material for more information on these datasets). The parameters and learned weights of source model $f_S(\cdot)$ were then used to initialise and train a new model on domain $\mathcal{D}_T$ and task $\mathcal{T}_T$ by transferring the source model layers and re-training (fine-tuning) the network towards this new target task $\mathcal{T}_T$ (i.e. in this case the subject group classification between HC, PwMSmild and PwMSmod). Figure \ref{fig:tl_schematic} schematically details the TL approach. Baseline ``end-to-end'' refers to a DCNN trained and validated exclusively on FL data. ``Direct''  transfer refers classification that is performed with full HAR trained model and weights; after, only the last fully connected dense layer has been replaced by disease targets and retrained on FL data (all other layers are frozen). ``Fixed'' transfer refer to a HAR trained architecture, where the convolutional blocks and weights are frozen and act as a ``fixed feature extractor'', however DNN weights thereafter are fine-tuned.\par
\subsubsection{Pre-Processing}
Several pre-processing steps were first performed to format the raw signals for input into the respective deep networks. To maintain consistency and for comparability using TL approaches, all signals were processed according to the same structure as \cite{RN988}. For additional consistency with \cite{RN988}, only subject's 2MWTs identified using the running belt were considered for subsequent analysis in this study. All inertial sensor signals were sampled at 50 Hz; in the case of the WISDM dataset, signals were resampled to 50 Hz using a shape-preserving piecewise cubic interpolation. Signals were filtered with  $4^{th}$ order Butterworth filter with a cut-off frequency at 17 Hz \cite{RN988}, and as per previous work, the smartphone coordinate axes were aligned with the global reference frame using the technique described in \cite{RN609}. 
All signals were detrended and amplitude normalised with zero mean and unit variance \cite{RN609}.
Sensor signals per each test were then up-sampled using fixed-width sliding windows of 2.56 sec and 50\% overlap (128 samples/epoch), in accordance with parameters in similar studies \cite{RN609,RN847, RN991}. The total number of observations/epochs for each constructed datasets are depicted in table \ref{table:constructed-dataset}.

\subsubsection{Model Evaluation}
 \begin{table}[t!]
\centering
\begin{footnotesize}
\centering
 \begin{threeparttable}
 \centering 
 \caption[\footnotesize{Overview of source $\mathcal{D}_S$ and target domain $\mathcal{D}_T$ datasets.}]{\footnotesize \textbf{Overview of source $\mathcal{D}_S$ and target domain $\mathcal{D}_T$ datasets}. Datasets were constructed from the  original sensor signal using 2.56 [s] epoch sliding windows with a 50\% overlap. }\label{table:constructed-dataset}
\begin{tabular}{l M{3cm} M{3cm} M{3cm}} \midrule\midrule
  & \multicolumn{2}{c}{$\mathcal{D}_S$}  & $\mathcal{D}_T$   \\\cmidrule(lr){2-3}\cmidrule(lr){4-4}
  $(n)$ & UCI HAR\tnote{1} & WISDM\tnote{1} & FL\\\midrule
   subjects & 30 & 51 & 97\tnote{a} \\
   tests & 61 & 252 & 970\tnote{b} \\
    samples & 10013 & 54781 & 82450\\\hline
\end{tabular}
\begin{tablenotes}\footnotesize
\item [1] See supplementary material for more information on the UCI HAR and WISDM datasets.  
\item[a] HC, n=24; PwMSmild, n=52; PwMSmod, n=21; see demographics table \ref{table:demographics} for more details.
\item[b] Randomly sampling $m=10$ tests per subject. 
\end{tablenotes}
\end{threeparttable}
\end{footnotesize}
\end{table}
To determine the generalisability of our models, stratified 5-fold, subject-wise, cross-validation (CV) was employed with the same seeding as in \cite{RN988}. This consisted of randomly partitioning the dataset into k=5 folds which was stratified with equal class proportions where possible. One set was denoted the training set (in-sample), which was further split for into smaller set for validation, using roughly 10\% of the training data proportionally. The remaining 20\% of the dataset was then denoted testing set (out-of-sample) on which predictions were made. \par
To help alleviate model biases occurring from the varying number of repeated tests contributed per subject over the duration of the FL study, $m$ number of 2MWTs per subject were randomly selected with replacement to create balanced datasets. Parameterisation of the number of tests per subject was determined using a baseline DCNN prior to building all subsequent models within FL. The classification performance over varying data set sizes was examined by sampling $m=\{1, 5, 10, 25, 50\}$ daily tests sampled (with replacement) per subject. It was observed that there was minimal additional classification performance after $m=10$ 2MWTs sampled per subject across each binary task. Class distributions in the training and validation sets were then balanced using the re-sampling approach in \cite{RN988}. Imbalances in the HAR training and validation data were also countered using this approach. The total number of observations/epochs for each constructed datasets are depicted in table \ref{table:constructed-dataset}.
HAR model performance was reported based on the classification of individual epochs into the correct activity class for UCI HAR and WISDM. FL and TL performance was based on the majority prediction of all epochs over a 2MWT, \textit{test-wise}, where final \textit{subject-wise} classification results were computed though majority voting each aggregated individual 2MWT prediction per subject (see \cite{RN988}). \par
Multi-class classification metrics were reported as the 2MWT test-wise median and interquartile range over one CV, as well as the final subject-wise outcome for that CV (in the case of FL), using overall metrics such as the macro accuracy, macro F1-score (MF\textsubscript{1}) and Cohen’s kappa ($k$) statistic \cite{RN1055,RN1027}. 

\subsection{Layer-Wise Relevance Propagation}
\label{sec:LRP}
Layer Wise Relevance Propagation (LRP) back-propagates through a network to decompose the final output decision, $f(\bm{x})$ \cite{RN861, RN851}. Briefly, a trained model's activations, weights and biases are first obtained in a forward pass through the network. Secondly, during a backwards pass through the model, LRP attributes relevance to the individual input nodes, layer by layer. For example $R_{k}$ denotes the relevance for neuron $k$ in layer $^{(l+1)}$, and $R_{j\leftarrow k}$ defines the share of $R_{k}$ that is redistributed to neuron $j$ in layer $^{(l)}$. The fundamental concept underpinning LRP is that the conservation of relevance per layer such that: $\sum_{j}R_{j\leftarrow k}=R_{k}$ and $R_{j}$ = $\sum_{k}R_{j\leftarrow k}$. The conservation of total relevance per layer can can also be denoted as:  
\begin{equation}\label{eq:conserve_relevance}
\sum_{j}R_{j\leftarrow k}^{(l, l+1)} = R_{k}^{(l+1)}
\end{equation}
Propagation rules are implemented to withhold this conservation property. Considering a DNN model, $a_{k} = \phi\left(\sum_{j}a_{j}w_{jk}+b_{k}\right)$, which consists of $a_{j}$, the activations from the previous layer, and $w_{jk}$, $b_{k}$, the weight and bias parameters of the neuron. The function $\phi$ is a positive and monotonically increasing activation function. In case of a component-wise operating non-linear activation, e.g. a ReLU $(\forall j=k : x_k = \max(0, x_j))$ then $\forall j=k : R_j = R_k$, since the top layer relevance values $R_k$ only need to be attributed towards one single respective input $j$ for each output neuron $k$. The $\alpha\beta$-rule for LRP has been shown to work well at decomposing a model's decisions:
\begin{equation}\label{eq:lrp_alpha_beta_rule}
    R_{j}=\sum_{k}\left(\alpha\frac{a_{j}w_{jk}^{+}}{\sum_{j}a_{j}w_{jk}^{+}} -\beta\frac{a_{j}w_{jk}^{-}}{\sum_{j}a_{j}w_{jk}^{-}} \right)R_{k},
\end{equation}
where each term of the sum corresponds to a relevance propagation $R_{j\leftarrow k}$, where $()^+$ and $()^-$ denote the positive and negative parts respectively, and where the parameters $\alpha$ and $\beta$ are chosen subject to the constraints $\alpha - \beta =1$ and $\beta\geq0$.
The $\alpha_1\beta_0$-rule ($\alpha$=1, $\beta$=0) emphasises the weights of positive contributions relative to inhibitory contributions predicting $f(\bm{x})$ and has been shown to create crisp and interpretable heatmaps in image recognition tasks \cite{RN851} and for explaining the presence of Alzheimer’s disease (AD) detection based on MRI imaging \cite{RN826}. For this first interpretation of LRP gait heatmaps we used $\alpha_1\beta_0$ to focus on the morphology and characteristics of a sensor signal influencing $f(\bm{x})>0$ with respect to that prediction. To benefit interpretation of LRP examples, we have standardised the heatmap colors, where hot hues represented to presence of factors which influenced $f(\bm{x})>0$, to predict MS disease, whereas cold hues (inversely) contradicted the prediction of MS (i.e. $f(\bm{x})<0$, or HC). A signed small stabilising term can be added to the denominator (termed $\epsilon$-rule):
\begin{equation}\label{eq:lrp_alpha_epsilon_rule}
    R_{j}=\sum_{k}\frac{a_{j}w_{jk}}{\epsilon + \sum_{j}a_{j}w_{jk}}R_{k},
\end{equation}
The $\epsilon$-rule has been shown to filter noisy heatmaps by absorbing some relevance when the contributions to the activation of neuron $k$ are weak or contradictory \cite{RN861, RN971}. For larger values of $\epsilon$ only the most prominent explanation factors are retained, yielding a more sparse and less noisy explanation. In accordance with \cite{ RN861}, $\alpha\beta$-rules were applied to convolutional layers and the $\epsilon$-rule ($\epsilon=0.01$) to dense layers. \par
In this study, individual LRP heatmaps were produced for epochs in the out-of-sample testing set using the iNNvestigate toolbox \cite{RN856}. For more information on the theoretical and practical implementation of LRP, we direct the reader to \cite{RN1225, RN856, RN861}. Both Keras and PyTorch implementations of the LRP algorithm have been developed and can be found at \url{http://heatmapping.org/}.

\subsection{Constructing Representative Gait Signal Epochs}
In order to determine the population-wise gait characteristics pertinent of healthy versus mild and moderate MS disease, an average representative epoch was generated for each subject-group using Dynamic Time Warping Barycenter Averaging (DBA) \cite{RN1203}. First, Dynamic Time Warping (DTW) is a method to measure the similarity (distance) between two sequences in cases where the order of elements in the sequences must be considered \cite{RN1204}. DTW can be used to align the signals such that the similarity between their points is minimised, hence generating a ``warped'' optimal alignment between sequences. For instance, gait cycle templates have previously been generated for PwMS using DTW \cite{RN618}.  
DTW Barycenter Averaging is a global averaging method for an arbitrary set of DTW sequences, which can be used to create a macro-average sequence, in this case, a representative gait epoch. An average gait cycle epoch for each subject-group was constructed by applying DBA to a random selection of correctly classified epochs (n=2000) with a high posterior probability of that class (Pr.$>$0.85), with no more than (k$<40$) epochs (i.e. $<50$\%) contributed from a single 2MWT. A previously trained DCNN model was then applied to each representative epoch and relevance scores were attributed using LRP. An implementation of DBA can be found at \url{https://github.com/fpetitjean/DBA/}. 

\section*{Acknowledgements}
The authors would like to thank all staff and participants involved in capturing test data. This study was sponsored by F. Hoffmann-La Roche Ltd. This research was supported by the National Institute for Health Research (NIHR) Oxford Biomedical Research Centre (BRC). This research also received funding from the Flemish Government under the ``Onderzoeksprogramma Artifici\"{e}le Intelligentie (AI) Vlaanderen'' programme.

\section*{Competing Interests}
During the completion of this work, A. P. Creagh was a Ph.D. student at the University of Oxford and acknowledges the support of F. Hoffmann-La Roche Ltd.; F. Lipsmeier is an employee of F. Hoffmann-La Roche Ltd; M. Lindemann is a consultant for F. Hoffmann-La Roche Ltd. via Inovigate; M. De Vos has nothing to disclose.

\section*{Data Availability Statement}
Qualified researchers may request access to individual patient level data through the clinical study data request platform (\url{https://vivli.org/}). Further details on Roche’s criteria for eligible studies are available here (\url{https://vivli.org/members/ourmembers/}). For further details on Roche’s Global Policy on the Sharing of Clinical Information and how to request access to related clinical study documents, see here (\url{https://www.roche.com/research_and_development/who_we_are_how_we_work/clinical_trials/our_commitment_to_data_sharing.htm}).

\section*{Code Availability}
Deep networks were built using Python v3.7.4. through a Keras framework v2.2.4 with a Tensorflow v1.14 back-end. Layer-wise Relevance Propagation (LRP) heatmaps were build using the iNNvestigate toolbox: \url{https://github.com/albermax/innvestigate}, that has been developed as part of the \url{http://heatmapping.org/} project. Visualisations were created using MATLAB v2019b. The Dynamic Time Warping Barycenter Averaging (DBA) methodology for creating average gait epochs was implemented using the code described at \url{https://github.com/fpetitjean/DBA/}. Experimental code can be found at: \url{https://github.com/apcreagh/MS-GAIT_InterpretableDL}.

\section*{Author Contributions}
\textbf{AC}: Conceptualisation, Methodology, Software, Formal Analysis, Writing - original draft, Writing - review \& editing; \textbf{FL}: Conceptualisation, Methodology, Writing - review \& editing, Supervision; \textbf{ML}: Conceptualisation, Methodology, Writing - review \& editing, Supervision;
\textbf{MDV}: Conceptualisation, Methodology, Writing - review \& editing, Supervision.

\printbibliography

\end{document}